\documentclass[authoryear,12pt]{elsarticle}

\makeatletter
\def\ps@pprintTitle{%
\let\@oddhead\@empty
\let\@evenhead\@empty
\def\@oddfoot{}%
\let\@evenfoot\@oddfoot}
\makeatother

\usepackage{amssymb}
\usepackage{times}  
\usepackage{helvet}  
\usepackage{courier}  
\usepackage[hyphens]{url}  
\usepackage{graphicx} 
\usepackage{natbib}  
\usepackage{caption} 
\DeclareCaptionStyle{ruled}{labelfont=normalfont,labelsep=colon,strut=off} 
\usepackage{algorithm}
\usepackage{algorithmic}

\usepackage[utf8]{inputenc} 
\usepackage[T1]{fontenc}    
\usepackage{hyperref}       
\usepackage{url}            
\usepackage{booktabs}       
\usepackage{amsfonts}       
\usepackage{nicefrac}       
\usepackage{microtype}      
\usepackage{xcolor}         

\usepackage{soul}
\usepackage{bm}
\usepackage{subfigure}
\usepackage{wrapfig}
\usepackage{amsmath}

\usepackage{diagbox}
\usepackage{multirow}
\usepackage{apalike}
\graphicspath{{Figures/}}

\newcommand{\vz}{\mathbf{z}}
\newcommand{\tvs}{\hat{\mathbf{s}}}
\newcommand{\vs}{\mathbf{s}}
\newcommand{\va}{a}
\newcommand{\fType}[1]{\textsf{#1}}
\newcommand{\aType}[1]{\texttt{#1}}
\newcommand{\actions}{\mathbf{A}}
\newcommand{\vu}{\mathbf{u}}

\newcommand{\hankz}[1]{{#1}}

\newcommand{\ignore}[1]{{}}

\newtheorem{remark}{Remark}

 \usepackage{indentfirst}
\journal{Expert Systems with Applications}

\begin{document}

\begin{frontmatter}



\title{Learning Visual Planning Models from Partially Observed Images}


\affiliation[inst1]{organization={School of Computer Science and Engineering, Sun Yat-sen University},
            city={Guangzhou},
            postcode={510006}, 
            state={Guangdong},
            country={China}}
\affiliation[inst2]{organization={School of Computer Science, Guangdong Polytechnic Normal University},            
            city={Guangzhou},
            postcode={510665}, 
            state={Guangdong},
            country={China}}
            
\author[inst1]{Kebing Jin}
\ead{jinkb@mail2.sysu.edu.cn}
\author[inst2]{Zhanhao Xiao\corref{cor1}}
\ead{xiaozhanhao@gpnu.edu.cn}
\author[inst1]{Hankui Hankz Zhuo\corref{cor1}}
\ead{zhuohank@mail.sysu.edu.cn}
\author[inst1]{Hai Wan}
\ead{wanhai@mail.sysu.edu.cn}
\author[inst1]{Jiaran Cai}
\ead{caijr5@mail2.sysu.edu.cn}
\cortext[cor1]{Corresponding Author} 

\begin{abstract}
{\color{black}
There has been increasing attention on planning model learning in classical planning. Most existing approaches, however, focus on learning planning models from structured data in symbolic representations. It is often difficult to obtain such structured data in real-world scenarios. Although a number of approaches have been developed for learning planning models from fully observed unstructured data (e.g., images), in many scenarios raw observations are often incomplete. In this paper, we provide a novel framework, \aType{Recplan}, for learning a transition model from partially observed raw image traces. More specifically, by considering the preceding and subsequent images in a trace, we learn the latent state representations of raw observations and then build a transition model based on such representations. Additionally, we propose a neural-network based approach to learn a heuristic model that estimates the distance towards a given goal observation. Based on the learned transition model and heuristic model, we implement a classical planner for images. We exhibit empirically that our approach is more effective than a state-of-the-art approach of learning visual planning models in the environment with incomplete observations.}
\end{abstract}




\end{frontmatter}

\section{Introduction}
Domain independent classical planning has been applied in a growing number of real-world scenarios.
It requires planning models to describe how the environment is changed by actions and in what \hankz{conditions} an action can be executed \cite{DBLP:journals/ai/JinZXWK22}. \hankz{Since creating planning models by hand is often time-consuming and resource-demanding, even for experts, automatically learning planning models from data has drawn increasing attention from researchers \citep{DBLP:journals/ai/AinetoCO19}}.
Most existing approaches focus on learning planning models from data in structured and symbolic representations \hankz{\citep{DBLP:journals/ai/YangWJ07,DBLP:journals/ai/Zhuo014,DBLP:journals/ai/ZhuoM014}}, such as Planning Domain Definition Language (PDDL), in order to take advantage of off-the-shelf efficient planners to compute plans.
In many real-world scenarios, such as camera-based security monitoring, it is often difficult to acquire structured data for learning planning models, since the world states are described by images that are unstructured (i.e., represented by pixels). It is challenging to learn planning models from unstructured raw data and conduct planning based on unstructured initial states and goal states since we need to consider the large feature space of raw data \citep{DBLP:conf/ijcai/AsaiM20}.

\begin{figure}
  \centering
  \includegraphics[width=0.5\textwidth]{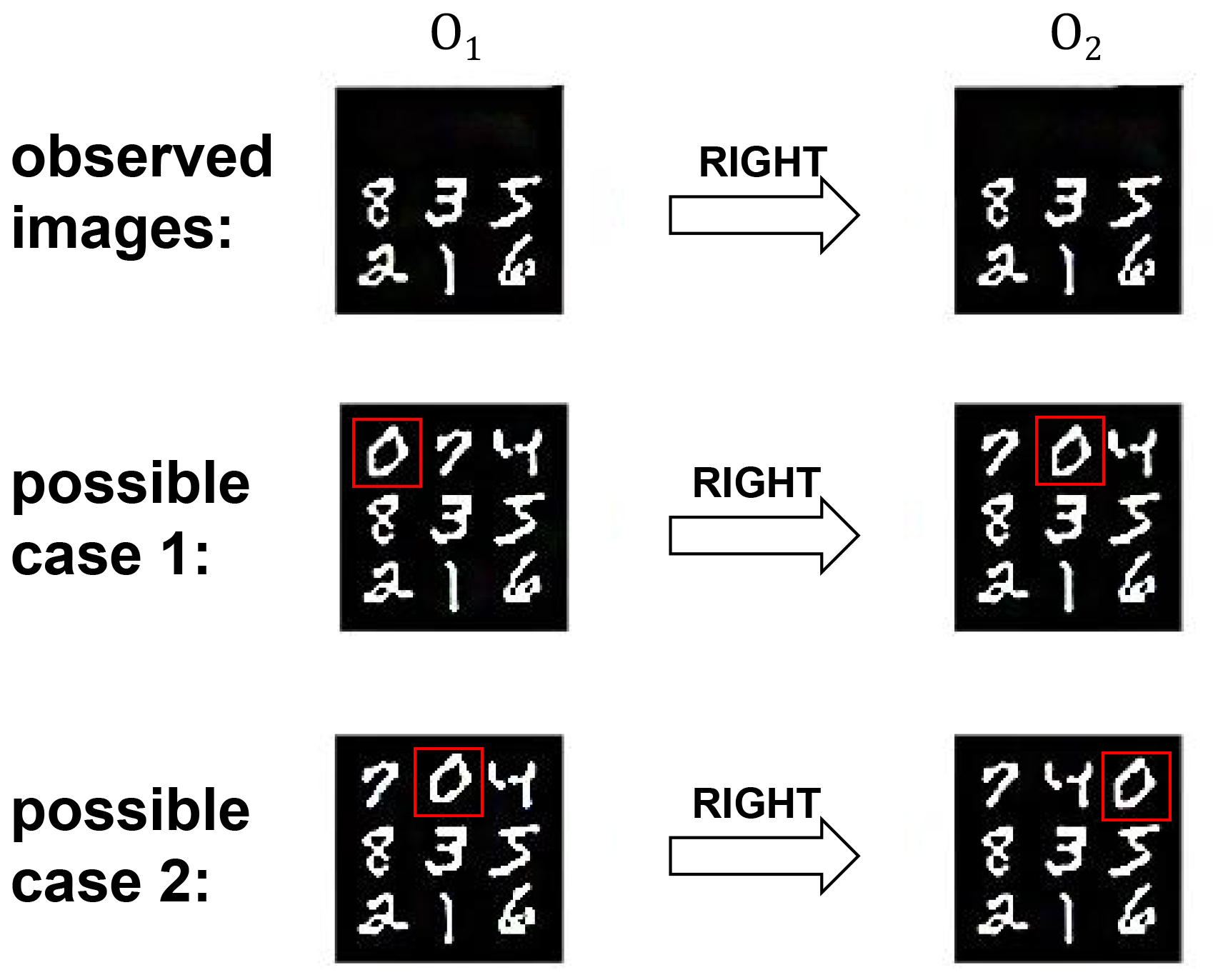}
  \caption{A transition pair of masked images in 8-puzzle domain. In each configuration, only the tile ``0'' can be exchanged with one of its adjacent tiles. An image observation $o_1$ is transformed into another image observation $o_2$ by executing action ``RIGHT''. When the first row is masked, there are at least two possible cases satisfying such observations.}\label{example_motivation}
\end{figure}

\hankz{Recently, \cite{DBLP:conf/aaai/AsaiF18} proposed a neural-symbolic approach, called \aType{Latplan}, for domain-independent image-based classical planning.
It learns latent (propositional or continuous) representations of images from a set of fully observed image transition pairs, 
and a transition function based on the learnt representations.}
\aType{Latplan} is applicable under the condition that each world state is fully observed in the form of an image, which in many real-world scenarios cannot be guaranteed.
On the contrary, it is more common that the world states in the form of images are partially observed. 
For example, an object may be occluded by an obstacle from the view of the camera, which results in images with partial observation.
Partial observation means that we need to handle the uncertainty of the world states, i.e., there exist different possibilities in the unobserved regions, \hankz{which makes the model learning task even more challenging.}
While \aType{Latplan} focuses on image transition pairs, it is sometimes impossible to entail the real information which is \hankz{missing}.
Figure \ref{example_motivation} shows an example in the 8-puzzle domain \hankz{to illustrate the motivation}.
There are a number of possible cases satisfying the transition pair when the three tiles in the first row are missing, two of which are shown in the second and third levels in Figure \ref{example_motivation}.
It is difficult to determine which of the two possible cases is true by only looking at the transition pair of the partially observed images, as \aType{Latplan} does. Whereas, these possible cases may be implied by having some future image observation. It suggests to handle the learning task in the perspective of a whole trace rather than a transition pair.  

Furthermore, the data is neither symbolic nor structured, which implies that symbolic rules are hardly likely to be learned without human knowledge. 
That is, it cannot be conjectured that the missing region only includes three numbers: ``0'', ``4'' and ``7''.
It is thus challenging to correctly recover the unobserved information and learn a planning model simultaneously.


In classical planning, it is satisfied that parts of world states unrelated to action execution will always keep unchanged.
Taking the example in Figure \ref{example_motivation}, if we observed that ``4'' occurs in the top right corner of the third image observation and the next action is ``DOWN'', we could conjecture that ``4'' also stays there in $o_2$, excluding the possible case 2.
In other words, to rule out possibilities, we are supposed to take advantage of the information about the previous and subsequent image observations. It thus motivates us to explore the underlying relations due to actions to help reason about the missing parts by taking a whole trace of image observations into account.

In this paper, we propose a framework \aType{Recplan} based on recurrent neural network (RNN), aiming to recover the missing information in the image observation traces and to learn a transition function simultaneously. 
More specifically, we first learn a prediction model to estimate the unchanged pixels and then complement the missing regions in the original image observations.
Following that, we learn a bidirectional mapping between image observations and latent states via an RNN which takes a whole trace as input. 
Meanwhile, we train a transition model in terms of the latent states.
During the training phase, the transition model in RNN takes not only the current state but also the history information as input, in order to reduce the possibilities due to the failure of observation.
Furthermore, we present a neural-network-based approach of learning a heuristic model in terms of latent states in order to estimate the distance to the goal state.
By conducting experiments on the three domains, we show that our approach is more effective in learning visual planning models in the environment with incomplete observations, compared with \aType{Latplan}.
We also show that our heuristic model is helpful in computing optimal plans.


This paper is organized as follows. We first review previous work related to our approach in the section and then formulate the planning problem.
After that we present our framework \aType{Recplan} in detail and evaluate \aType{Recplan} in three domains with comparing to \aType{Latplan}.
Finally, we conclude our paper and address future work.

\section{Related Work}
\subsection{Visual Planning}
In order to handle goal-directed planning problems with raw observations, there have been efforts in visual planning recently. Some researchers focus on applying learning-based approaches to robotic manipulation directly from raw observations, while some pay attention on learning planning models and state representation from image traces.

Many approaches have borrowed deep learning methods to tackle complex visual planning problems, which aim at generating an action sequence to reach a given goal observation from a given initial observation. For example, \cite{DBLP:conf/icra/NairCAIAML17} proposed to learn a model to manipulate deformable objects. Recently, Causal InfoGAN \citep{DBLP:conf/nips/KurutachTYRA18} was proposed to learn deformable objects from sequential data and generate goal-directed trajectories. In 2019, \citeauthor{DBLP:conf/icml/HafnerLFVHLD19} gave a planning framework PlaNet that learns the environment
dynamics from images and chooses actions
through predicting the rewards ahead for
multiple time steps. 
To compute goal-directed plans, SPTM \citep{DBLP:conf/icml/LiuKTAT20} was proposed. It builds a connectivity graph where each observation is considered as a node. Besides, HVF \citep{DBLP:conf/iclr/NairF20} was presented to compute plans by decomposing a visual goal into a sequence of subgoals. However, the above approaches suppose a fully observable environment where each observation is completely obtained, which in fact is a demanding requirement.

Researchers also show their interests in representation learning from image sequences. First of all, Asai and Fukunaga proposed \aType{Latplan} \citep{DBLP:conf/aaai/AsaiF18} 
and opened the door to image-based domain-independent classical planning, which bridges the gap between deep learning perceptual systems and symbolic classical planners.
Later \cite{DBLP:conf/aips/Asai19} offered an approach to obtain first-order logic representation from images, which is plannable with human knowledge.
Recently, based on \aType{Latplan}, \cite{DBLP:conf/ijcai/AsaiM20} proposed to learn planning models that are neural network restricted to cube-like graphs, which increases significantly planning accuracy.
Compared with the series of works about \aType{Latplan}, we in this paper assume that images are partially observed.
In consequence, we require that action labels are given in the training data.
If we followed their unlabeled assumption, the learning task would become intractable as there is little information to reason.
Another difference lies in that we take the whole image traces as training data while they adopt a set of image transition pairs.

\subsection{Inpainting}
Our paper is also related to the work of image inpainting and video inpainting. 
Image inpainting is rooted in \citep{DBLP:conf/siggraph/BertalmioSCB00}, which aims to fill corrupted regions of images with fine-detailed contents.
Existing inpainting methods can be classified into two categories: classical approaches and deep learning based approaches.
The classical approaches either replace incomplete regions with surrounding textures and apply a diffusive process \citep{DBLP:conf/cvpr/RothB05}, or fill holes by searching similar patches from the same image or external image databases \citep{DBLP:conf/eccv/He012}.
However, such kinds of approaches fail to capture semantical content and only work on the images with simple and repeated textures.
The recent success of deep learning approaches has irradiated such a conventional image process task and has inspired a number of deep learning based approaches.
For example, the first attempt is Context Encoder \citep{DBLP:conf/cvpr/PathakKDDE16} which uses a deep convolution encoder-decoder.
It has inspired a series of work that extend Context Encoder in different ways, such as \citep{DBLP:journals/tog/IizukaS017}, \citep{DBLP:conf/eccv/YanLLZS18}, etc.
Recently, researchers have paid more attention on exploiting image structure knowledge for inpainting \citep{DBLP:conf/cvpr/XiongYLYLBL19,DBLP:conf/aaai/YangQS20}.
However, existing image inpainting approaches only focus on an image itself without analyzing images on a transition pair or a trace.
It leads that these approaches still suffer the issue in Figure \ref{example_motivation}, failing to eliminate possible cases.

\subsection{Planning Model Learning}
Planning model learning has been attracted a lot of attention and there exist a number of approaches \citep{DBLP:journals/ker/AroraFPMP18}.
The seminal learning approach from partially observed plan traces should be ARMS~\citep{DBLP:journals/ai/YangWJ07}, which invokes a MAX-SAT solver to obtain planning models.
It has inspired a series of learning approaches \citep{DBLP:conf/ijcai/ZhuoK13,DBLP:conf/aaai/Zhuo15, DBLP:journals/ai/ZhuoYHL10, DBLP:journals/ai/ZhuoK17}, etc.
Besides, there is a large body of work from partially observed plan traces (LOCM~\citep{DBLP:journals/ker/CresswellMW13}, NLOCM~\citep{DBLP:conf/aips/GregoryL16}), PELA~\citep{DBLP:conf/aips/0004ATRI16}), from state pairs (FAMA \citep{DBLP:journals/ai/AinetoCO19})
and from labeled graphs \citep{DBLP:conf/ecai/BonetG20}.
Whereas, the above approaches require to structured symbolic training data, which leads a difficult and time-comsuming task.

\subsection{Heuristics Learning in Classical Planning}
Our paper also relates to the work of learning heuristics in classical planning via machine learning techniques \citep{DBLP:conf/socs/ArfaeeZH10,DBLP:conf/aips/ShenTT20,DBLP:conf/icaart/TrundaB20,DBLP:conf/ecai/FerberH020}.
It is also related to learning control knowledge \citep{DBLP:journals/jmlr/YoonFG08} and learning search policies \citep{DBLP:conf/aips/GomoluchARB20} in classical planning.
These approaches are applicable to descriptive planning models.
While we learn a transition model on the latent state of images, these approaches become inapplicable.
On the other hand, Asai and Muise (\citeyear{DBLP:conf/ijcai/AsaiM20}) recently presented to learn descriptive planning models from images, which allows applying state-of-the-art heuristics planners.
However, the approach is still only applicable to fully observed images.

\section{Transition Model Learning}
In this section, we first formulate the problem of transition model learning and then present our framework \aType{Recplan} that includes: (i) an inpainting model to complete masked image observations, (ii) a State Autoencoder (SAE) to embed a completed image into a latent propositional state, (iii) a transition model to update a latent state by an action.

\subsection{Problem Formulation}
Let $O^i = \langle o^i_0,o^i_1,...,o^i_n \rangle$ be a sequence of image observations, $A^i = \langle a^i_1,a^i_2,...,a^i_n \rangle$ be a sequence of actions and $M^i = \langle m^i_0,m^i_1,...,m^i_n \rangle$ be a sequence of mask matrices with the same size of image observations. The superscript $i$ is an identifier of a sequence and the subscript indexes elements in a sequence. Intuitively, the image observation $o^i_j$ is resulted from executing the action $a^i_j$ on the image observation $o^i_{j-1}$, according to an underlying transition function. In every mask matrix, value ``1'' denotes an observed pixel while value ``0'' denotes a masked one.

We define the image transition model learning problem as a tuple $(\{O^i\},\{M^i\},\{A^i\})$, which aims to learn a transition model $\hat{\gamma}$ to approximate the ground-truth transition function $\gamma$.
Formally, given any unmasked image $o$ and an applicable action $a$, $\hat{\gamma}(o,a) \approx \gamma(o,a)$.

\begin{remark}
Given the success of occlusion detection approaches \citep{Suresh2013ASO,DBLP:journals/ijaip/AskarERYE20}, which can be used to detect the masks in the given images, we believe it is reasonable to incorporate mask matrices as an input of the learning problem. In other words, it is not difficult to obtain mask matrices of raw observations from a realistic perspective.
\end{remark}

\subsection{Observation Inpainting}
\hankz{Considering there exist strong relations between each two continuous image observations, i.e., the latter image observation is the result of applying an action to the former image observation, we aim to exploit those relations to help ``complement" the missing parts of the image observations.}
In classical planning, only direct effects are considered: each pixel in the image keeps unchanged until it is affected by actions.
Based on such an assumption, we propose to learn a prediction function $\rho$ that takes two neighbor image observations $o^i_{j-1},o^i_{j}$ in a sequence and the connecting action $a^i_j$ as input, and outputs a matrix $\hat{p}^i_j$ with the same size of the image observation. Every element in $\hat{p}^i_j$ is a real number between $-1$ and $1$, which means to be a confidence on whether the corresponding pixel in $o^i_{j}$ changes or not.
When it is closer to $1$, the corresponding pixel will be more likely to be unchanged in the original image of the observation $o^i_j$.

Indubitably, it is a semi-supervised task: the pixels in each image observation which are observed to be unchanged are labeled as unchanged; those unobserved pixels are unknown.
Then we construct the training data as a set of label matrix sequences $\{\langle p^i_1,p^i_2,...,p^i_n \rangle\}$.
Formally, for an element $(x,y)$ in a label matrix $p^i_j$, denoted by $p^i_j(x,y)$, its value is defined as follows:
\begin{itemize}
\item If $o^i_{j-1}(x,y) = o^i_{j}(x,y),\ m^i_{j}(x,y)=m^i_{j-1}(x,y)=1$, then $p^i_j(x,y) =1$;
\item if $o^i_{j-1}(x,y) \neq o^i_{j}(x,y),\ m^i_{j}(x,y)=m^i_{j-1}(x,y)=1$, then $p^i_j(x,y) =-1$;
\item otherwise $p^i_j(x,y) =0$.
\end{itemize}
Intuitively, when a pixel $(x,y)$ in $o^i_j$ is explicitly observed to be unchanged, it is labeled as ``$1$'' in the label matrix $p^i_j$ and is labeled as ``$-1$'' when it is definitely observed to be changed. When it is masked in the either current or the next observation, it is labeled as ``$0$'', with the meaning of unknown.


The objective of the training phrase is to learn a prediction function that exploits the relations between image observations caused by actions. When it converges, for each image observation $o^i_j$, we can obtain a matrix that consists of the prediction result on each pixel and we use $\hat{p}^i_j$ to denote it.


Next, we complement the image observations based on the prediction matrices.
Given an image observation sequence $O^i$, a mask matrix sequence $M^i$ and a prediction matrix sequence $\hat{P}^i$, we define the complemented image sequence $\overline{O}^i = \langle \overline{o}^i_1,\overline{o}^i_2,...,\overline{o}^i_n \rangle$ as follows: for every pixel $(x,y)$,
\begin{align*}
\overline{o}^i_{j-1}(x,y) =
\begin{cases}
{o}^i_{j}(x,y) & \mbox{if } m^i_{j-1}(x,y){=} 0, m^i_{j}(x,y) {= }1,\\
& \phantom{if } \hat{p}^i_{j}(x,y) > \lambda\\
o^i_{j-1}(x,y)  & \mbox{otherwise}
\end{cases}\end{align*}
where $\lambda$ is a predefined positive threshold.

\begin{figure}
    \centering
    \includegraphics[width=0.8\textwidth]{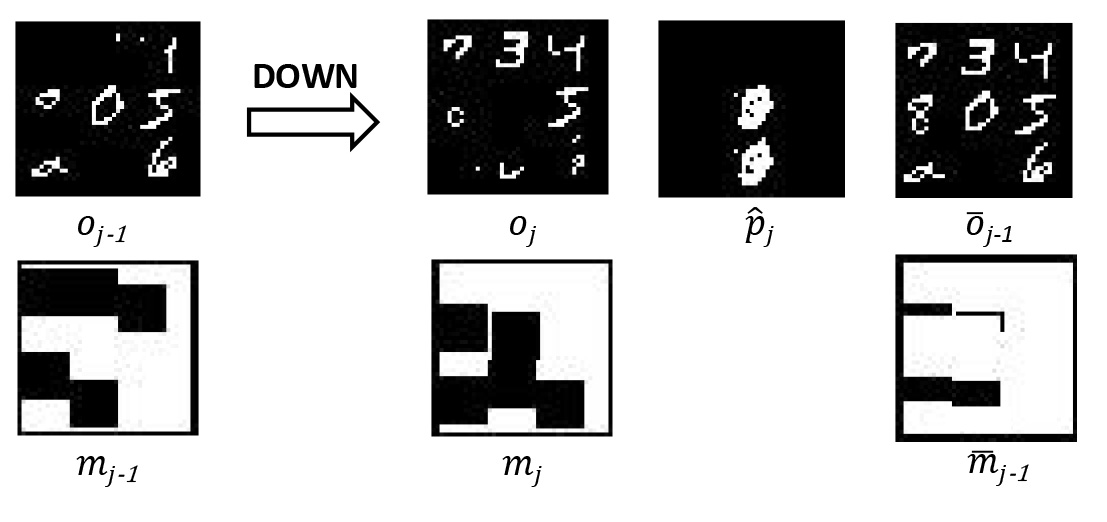}
    \caption{An example of complementing $o_{j-1}$ with $o_j$}
    \label{fig:inpainting_example}
    \setlength{\belowcaptionskip}{-2cm} 
\end{figure}

Intuitively, when the confidence for a masked pixel is higher than the threshold $\lambda$, it is predicted as unchanged. 
It can be substituted with an unmasked pixel in the latter image observation which is predicted as unchanged. 
Meanwhile, we update the mask matrices for the complemented pixels, denoted by $\overline{M}^i = \langle \overline{m}^i_1,\overline{m}^i_2,...,\overline{m}^i_n \rangle$. Formally, if ${o}^i_{j}(x,y) \neq \overline{o}^i_{j}(x,y)$, we set $\overline{m}^i_j(x,y) = 1$.


We use an example in Figure \ref{fig:inpainting_example} to show how to complement $o_{j-1}$ with $o_j$.
White pixels in the prediction matrix $\hat{p}_j$ are predicted as changed. Then we obtain a complemented image $\overline{o}_{j-1}$ by replacing with the unmasked pixels in $o_j$ that are labeled as unchanged by $\hat{p}_j$ and update the mask matrix $m_{j-1}$ to $\overline{m}_{j-1}$.


\begin{remark}
It is notable that the above inpainting approach adopts a complementing procedure from back to front.
One may argue that by considering the complementing procedure in dual directions, more masked regions could be complemented. However, the neural network-based prediction model cannot guarantee perfect correctness, and such a bidirectional procedure may result in a contradiction. In other words, it possibly happens that a masked pixel is predicted as unchanged in two neighbour images but it would be complemented as different pixels.   
\end{remark}

\subsection{Representation Learning}
Inspired by \aType{Latplan}, we introduce the State Autoencoder (SAE) of our framework \aType{Recplan}.
SAE is a variational autoencoder~(VAE) neural network architecture \citep{DBLP:conf/nips/KingmaMRW14} with a Gumbel-Softmax activation \citep{DBLP:conf/iclr/JangGP17} which aims to learn a bidirectional mapping from image observations to latent states. SAE contains two neural networks: one is called \fType{Encoder}, translating an image observation into a latent state and the other one is called \fType{Decoder}, reconstructing an image from a latent state. 

\begin{figure}[!ht]
  \centering
  \includegraphics[width=0.8\textwidth]{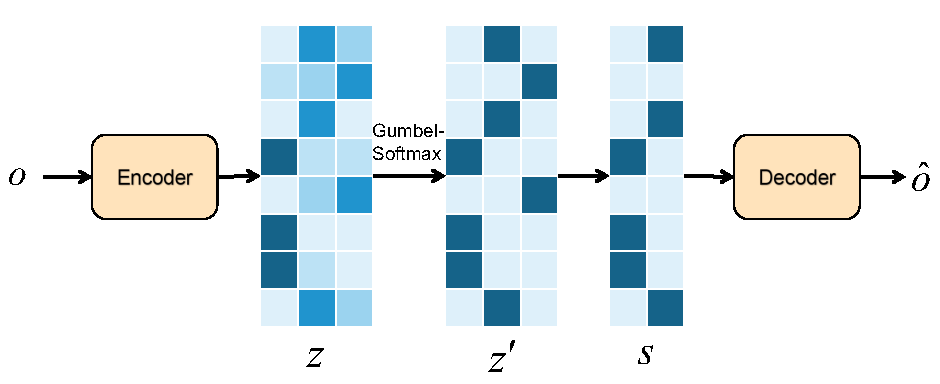}
  \caption{The procedure of the State Autoencoder}\label{fig_embedding}
     \setlength{\belowcaptionskip}{-2cm} 
\end{figure}

The whole procedure of SAE is shown in the Figure \ref{fig_embedding}. First, an image observation $o$ is translated into a $k\times 3$-dimension matrix $\vz$ via \fType{Encoder}. We further translate $\vz$ into a $k\times 3$-dimension matrix $\vz'$ via a Gumbel-Softmax activation function. The Gumbel-Softmax activation function is a reparametrization trick which connects the continuous space and the discrete binary space.
Every row in $\vz'$ is a one-hot representation of three exclusive cases: ``True'', ``False'', and ``Unknown''. 
More specifically, $(1,0,0)$ indicates a bit is ``True'', $(0,1,0)$ indicates ``False'', and $(0,0,1)$ indicates ``Unknown''.
We call the output matrix $\vz'$ a latent state and divide it into two parts: a matrix composed of ``True'' and ``False'' columns, denoted by $\vs$, and the ``Unknown'' column, denoted by $\vu$.
Intuitively, each bit of a latent state potentially represents some region of an image, which is implemented via \fType{Decoder}.
Actually, the ``True/False'' matrix itself is sufficient to represent a latent state because ``(0,0)'' means to be ``Unknown''. So we sometimes simply call the matrix a latent state as well. 

The objective of training \fType{Encoder} and \fType{Decoder} is to make the original image observation $o$ and the reconstructed image $\hat{o}$ become identical as possible in those observed regions. Then we employ the Mean Square Error (MSE) as the reconstruction loss function:
\begin{align*}
L_{recon}= \Sigma_i \Sigma_j ||o^i_j \odot m^i_j  - \hat{o}^i_j \odot m^i_j||^2
\end{align*}
where $\odot$ is element-wise multiplication.

To obtain a more accurate latent state, we design a loss function based on the relation between the image observation and its complemented image.
As the complemented image $\overline{o}$ differs from the image observation $o$ only on the masked regions, except for the ``Unknown'' rows, the latent state $\overline{\vs}$ of $\overline{o}$ should coincide with $\vs$ of $o$.
On the other hand, the image observation has more masks than the complemented image. Thus, the ``Unknown'' bits of $\vu$ should include those of $\overline{\vu}$.
Then we have the following loss function for latent states:
\begin{align*}
L_{state} = \Sigma_i\Sigma_j(||\overline{\vs}^i_j\odot\vs^i_j - \vs^i_j||^2 + ||\overline{\vu}^i_j\odot\vu^i_j- \vu^{i}_j||^2).
\end{align*}

Finally, we define the loss function of SAE as:  
\begin{align*}
L = L_{recon} + L_{state}.
\end{align*}

\subsection{The Learning Procedure}
\begin{figure}[!ht]
  \centering
  \includegraphics[width=0.9\textwidth]{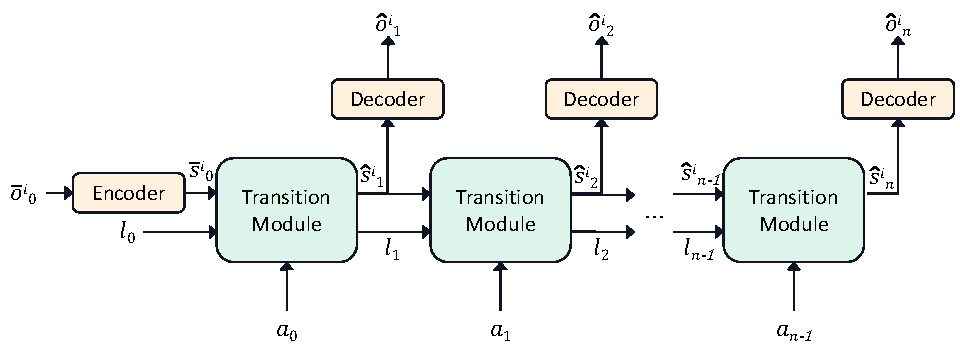}
  \caption{The transition model learning procedure of \aType{Recplan}}
  \label{fig_rnn}
\end{figure}

Next, we learn a transition model to simulate the underlying transition function that propagates an image into another image according to some action. 
Different from \aType{Latplan} which only takes image transition pairs without action labels as input, we propose to learn a transition model via a recurrent neural network (RNN) in order to capture the unobserved information from a successive image observation trace. Figure \ref{fig_rnn} shows \hankz{the learning procedure of the transition model}. In the beginning, the transition module takes as input the latent state $\overline{\vs}^i_0$ of the initial complemented image $\overline{o}^i_0$, the first action $\va^i_0$ and the initial history information $l_0$. At every following step, the transition module takes the latent state $\vs^i_{j}$ and the history information $l_j$ obtained from the previous step and an executed action $\va^i_j$. Then the transition module outputs a latent state $\vs^i_{j+1}$ which can be decoded as an image $\hat{o}^i_{j+1}$ and a history information $l_{j+1}$ to the next step. Here actions are denoted by vectors in the one-hot representation.

Intuitively, the next latent state computed by the transition module is restricted by the reconstruction loss not only from the current image, but also from the previous images.

To approximate the underlying transition function, we require the image $\hat{o}^i_{j}$ decoded from $\vs^i_{j}$ to coincide with the complemented image $\overline{o}^i_{j}$ on the observed regions.
We thus use MSE as the loss function:
\begin{align*}
L_\gamma = \Sigma_{i}\Sigma_j ||\overline{o}^i_j - \hat{o}^i_j \odot \overline{m}^i_j ||^2
\end{align*}

When the loss function converges, every image observation is recovered to a complete image and
the initial history information $l_0$ is fixed.
More specifically, given any action $a$ and any image $o$, we first translate $o$ into a latent state $\vs$ via \fType{Encoder}, and 
compute its next latent state $\vs'$ via the transition module with $\vs$ and $\va$ as input, and then 
apply \fType{Decoder} to reconstruct an image $o'$ for $\vs'$.
That is, we obtain a transition model $\hat{\gamma}$ such that $o' = \hat{\gamma}(o,a)$, which recovers the masked images as much as possible in the input image observation traces.

\section{Visual Planning and Heuristic Model Learning}
In this section, we show how to learn a heuristic function based on latent states and propose a goal-oriented planning approach for images.
\subsection{Heuristic Model Learning}
Essentially, latent states represent images in a high dimensional space. It actually provides a way to exploit relations among images and actions from another perspective. 
For that, we propose to learn a neural-network-based heuristic function based on latent states.
Basically, we consider the heuristic learning problem as a regression problem which aims to find a value that describes the distance from the current image to the goal image in an image sequence.

First, we show how to construct the training data for this learning task.
We define the objective heuristic value for the latent state $\tvs^i_{j+1}$ as the number of actions to obtain the goal image $o^i_n$ from the current image $o^i_j$. Formally, the heuristic value of selecting the action $\va^i_{j+1}$ on the latent state $\tvs^i_{j}$ is $n-j-1$. 
In order to select the smallest heuristic value at each step, we hope that the heuristic value of the appropriate action is smaller than that of other actions. Thus, we simply set the heuristic value as $n-j+1$ when selecting other actions rather than $\va^i_{j+1}$ on the latent state $\tvs^i_{j}$. Formally, for an image observation sequence ${O}^i$ with its goal image $o^i_n$ and every action $a$ occurring in all sequences, we define the training data about heuristic values as follows: for $0 \leq j < n$,
\begin{align*}
\label{eqn_loss_heuristics_label}
h(\tvs^i_j,\overline{s}^i_n,\va) =
\begin{cases}
    n-j-1, & \mbox{if } \va = \va^i_{j+1} \\
    n-j+1, & \mbox{otherwise}
  \end{cases}
\end{align*}
where $\tvs^i_j$ is the latent state of $o^i_j$ computed by $\aType{Recplan}$ and $\overline{s}^i_n$ is the latent state of $o^i_n$ computed by $\fType{Encoder}$.

\begin{remark}
Notably, the reason why we use $\overline{s}^i_n$ rather than the latent state of the complete image computed by $\aType{Recplan}$, i.e., $\tvs^i_n$, is because we hope to learn a heuristic model that is applicable to a goal image with masks. It allows the approach to conduct planning in a partially observable environment.
\end{remark}

The learning task is to obtain a heuristic model $\hat{h}$ to simulate the objective heuristic function $h$.
Finally, we take MSE as the loss function: 
\begin{align*}
L_h = \Sigma_i \Sigma_j||h(\tvs^i_j,\overline{s}^i_n,\va)-\hat{h}(\tvs^i_j,\overline{s}^i_n,\va)||^2.
\end{align*}


\subsection{Visual Planning}
The visual planning task aims to compute an action sequence leading a given initial image to propagate to be a given goal image and generate the intermediate images between the two given images.

\begin{figure}[!ht]
  \centering
  \includegraphics[width=0.9\textwidth]{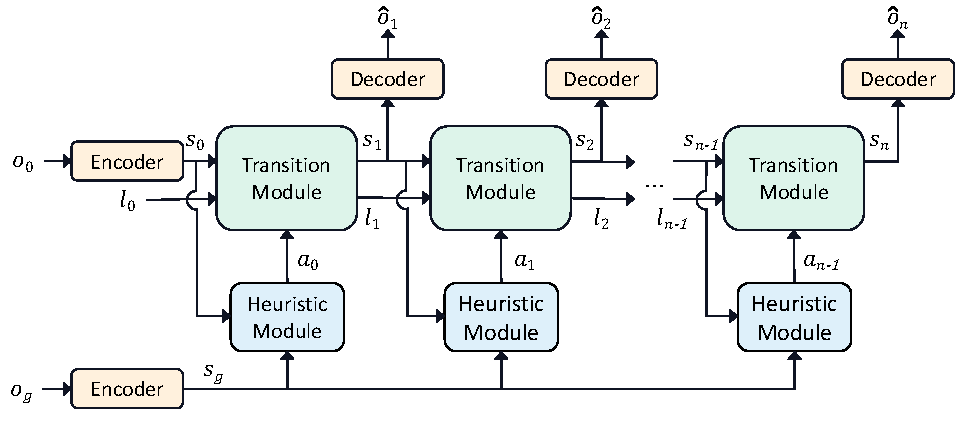}
  \caption{The planning procedure of \aType{Recplan}}\label{fig_planning}
\end{figure}

Indeed, we do not learn a declarative PDDL-like planning model, which leads it impossible to invoke an off-the-shelf PDDL-based planner. Similar to \citep{DBLP:conf/aaai/AsaiF18}, we implemented an $A^*$ algorithm for visual planning based on the learned transition model. But different from \aType{Latplan} which trivially takes the number of the unachieved goals as the heuristic value, we apply the learned heuristic model.

Figure \ref{fig_planning} illustrates the planning procedure of our framework $\aType{Recplan}$.
Different from the training phase where a whole image observation sequence is considered as input, in the planning phase
\aType{Recplan} takes as input a fully observed initial image $o_0$ and a possibly partially observed goal image $o_g$. Finally it outputs a plan with a sequence of complete images connecting the initial image $o_0$ and a complete image that coincide with $o_g$. 
At every step, the heuristic module takes the current latent state $\vs_j$ and the latent state $\vs_g$ of the goal image as input.
It outputs an action $\va_j$ with the smallest heuristic value via traversing all actions.
Meanwhile, the transition module has four inputs: the latent state $\vs_j$ which is computed from the previous step, and the action $\va_j$ obtained from the heuristic module and the history information $l_j$. It outputs a latent state $\vs_{j+1}$ which will be decoded into an image $\hat{o}_{j+1}$ and will be propagated to the next step.

When the decoded image $\hat{o}_j$ for some step is close to the goal image despite masks, i.e., their difference is smaller than a quite small threshold, the planning algorithm terminates.
The sequence of the actions output by the heuristic module is a solution plan and the sequence of decoded images describes how the initial image evolves to the goal image. 

\ignore{
The planning procedure is a heuristic searching algorithm.
Given an initial image $o_0$ and a goal image $o_g$, \aType{Recplan} first computes a latent state $s_0$ and $s_g$ by SAE \fType{Encoder}. 
Then for each action, we estimate a heuristic value $e$ by the learned heuristic model. 
We use $g$ to denote the length of the steps that the plan has taken, where $g=0$ in the beginning.  
At every step, the action $a$ which has the minimal estimation $f = g+e$ will be selected. 

Next we repeat the searching procedure until the goal image is reached (line 11-12) or running time exceeds the cutoff time $t_{max}$. 
We first find the node $(f,g,e,p,s)$ with the minimal $f$ in $queue$. 
If the decoded image of the last state in $S$ differs with the goal image within a threshold $\lambda'$, we terminate and return the plan and the corresponding image sequence.
If the goal state is not obtained, we update the state and increase $g$ by one. 
We repeatedly compute the heuristic value for each action and expand the queue.

\begin{algorithm}[H]
\caption{Visual Planning}
\label{al:heuristic}
\textbf{Input}:Initial image $o_0$ and goal image $o_g$\\
\textbf{output}:A plan $p$ and an image sequence
\begin{algorithmic}[1] 
\STATE $s_0$ = \fType{Encoder}($o_0$), $s_g$ = \fType{Encoder}($o_g$), $queue$ = []
\FOR{each $a$ in $\actions$}
\STATE $g = 0$, $e = h(s_0,s_g,a)$
\STATE $f = g + e$ 
\STATE $p = [a]$
\STATE $S = [s_0]$
\STATE $queue = [ queue|(f,g,e,p,S) ]$
\ENDFOR
\WHILE{$queue$ is non-empty and running time is less than $t_{max}$}
\STATE $(f,g,e,p,S) = queue[i]$ where $f$ of $queue[i]$ is minimal 
\STATE $s = last(S)$
\IF{$|\fType{Decode}(s) - o_g| \leq \lambda'$}
\STATE \textbf{return} $p$ and $\fType{Decode}(S)$
\ENDIF
\STATE $s' = \gamma(s,a)$
\STATE $g= g+1$
\FOR{each $a$ in $\actions$}
\STATE $e = h(s',s_g,a)$
\STATE $f = g+e$
\STATE $queue = [queue |  (f,g,e,[p|a],[S|s'])]$
\ENDFOR
\ENDWHILE
\end{algorithmic}
\end{algorithm}
}

\section{Experiments}
\subsection{Domains and Data Sets}
Next, we evaluate our approach \aType{Recplan} on three domains: 8-puzzle (MNIST) and 8-puzzle (Mandrill, Spider).

\noindent
\textbf{8-puzzle (MNIST).} Every configuration in an 8-puzzle domain is composed of 9 digits (0-8) which are arranged in a $3\times 3$ tile matrix. The tile ``0'' is considered as blank. 
The only valid actions are swapping the blank tile with one of its adjacent tiles. We use ``UP'', ``DOWN'', ``RIGHT'' and ``LEFT'' to denote the valid actions. For any configuration we generate a $42\times42$ image obtained by replacing each digit tile with the corresponding digit image from the MNIST data set \citep{lecun1998gradient}. 
%

\noindent
\textbf{8-puzzle (Mandrill and Spider).}
For these two domains, we divide two pictures Mandrill and Spider into nine sub-images by a $3 \times 3$ matrix, respectively. Each sub-image is labeled by nine digits (0-8).
The other settings are similar to the 8-puzzle (MNIST) domain.
Then each state yields a $42\times42$ image composed of nine muddled sub-images.

\ignore{
\noindent
\textbf{Blocks.}
The Blocks domain describes how to pick and place blocks on a table by a robot hand.
It involves four classes of actions: ``PICK-UP'' for picking up a block from a table, ``PUT-DOWN'' for putting down a block on a table, ``STACK'' for putting a block on another block and ``UNSTACK'' for lifting a block from another block.
In this domain, we consider every state includes five blocks, in consequence there are 35 grounding actions.
For each state, we apply the PDDLGym tool \citep{silver2020pddlgym} to generate a  $42 \times 42$ image.

\noindent
\textbf{Hanoi.}
The Hanoi domain describes the classical scheduling problem that aims to move disks with different sizes among several pegs. It only involves one class of actions, ``MOVE'' for moving a disk from one peg to another peg. Each movement needs to guarantee bigger disks are below smaller disks.
In this domain, we consider four disks and three pegs, so there are 18 grounding actions.
Similarly, we apply the PDDLGym tool to generate a  $32 \times 96$ image for every state.
}

It is remarkable that our heuristic learning approach does not require the training sequences to be optimal.
If valid plans rather than optimal plans are used as training data, it would lead that the learned heuristic model computes plans that are not optimal but valid.
However, in order to assess approaches in terms of plan quality, we set out to create optimal plans as data set.
First, for each domain, we randomly generate initial states. 

Following \citep{DBLP:conf/aaai/AsaiF18}, we use a breadth-first search algorithm to enumerate action sequences with a length of $7$ in order to construct a tree in which each node represents a state and each edge represents an action. We build the data set by selecting those plans whose final state never appears in the tree with a level less than $8$, which thus are assured to be optimal.
Then, for each domain, we generate 10100 distinct state-action sequences.
We take 8500 sequences for training, 1500 for validation and 100 for testing.

After creating an image observation for each state as we mentioned above, we randomly generate mask matrix sequences with different mask sizes and numbers.
By element-wise multiplication, we then obtain a set of image observation sequences.


\subsection{Experiment Details}
In this paper, we need to train five models. The followings are the details of the models:
\begin{itemize}
    \item  We construct the prediction model as a neural network composed of two 3$\times$3 convolutional layers, two 400-dimension fully connected layers and a tanh activation function. 
    \item  We construct the SAE \fType{Encoder} as a neural network with two 3$\times$3 convolutional layers with tanh as activation function, a 1000-dimension fully connected layer and a 72$\times$2-dimension output layer.
    \item The SAE \fType{Decoder} consists of two 1000-dimension fully connected layer, with ReLU as activation function and outputs a 42$\times$42 image.
    \item  We construct the transition model as 400-dimension RNN cell, a 1000-dimension fully connected layer for 8-puzzles domains and a 250-dimension fully connected layer for blocks domain and hanoi domain, a ReLU activation function and a 72-dimension output layer. 
    \item  The heuristic model requires to vary based on the number of actions.
    We construct the heuristic model by three 76-dimension fully connected layers and a 4-dimension output layer, with LeakyReLU as the activation function.
\end{itemize}

We train all models with batch size 500 (250 for Hanoi domain) and learning rate $10^{-3}$.
Specifically, we train
(1)~the prediction model for 500 epochs; 
(2)~the SAE \fType{Encoder} and \fType{Decoder} for 1000 epochs with Gumbel-Softmax temperature: 5.0 $\rightarrow$ 0.7, droptout factor 0.4 and Batch Normalization \citep{DBLP:conf/icml/IoffeS15};
(3)~the transition model and the heuristic model for 1500 epochs. 
All experiments are conducted on a machine with a 12GB GeForce GTX 1080 Ti.

{\color{black}\textbf{Baseline.}
In this paper, we take the seminal image-based classical planning approach \aType{Latplan} \citep{DBLP:conf/aaai/AsaiF18} as the baseline approach and we use \aType{Lat} to denote it.
As mentioned above, \aType{Latplan} is proposed for the fully observed setting, so we add the inpainting approach to predict the missing parts before training the networks of \aType{Latplan}.
Then we use \aType{Lat}$^+$ to denote it.
On the other hand, we use \aType{Rec} to denote our approach \aType{Recplan}.
Similarly, for comparision, we also equip our inpainting approach into \aType{Recplan} and use \aType{Rec}$^+$ to denote it.
}

\subsection{Metrics}

We evaluate the approaches on the following aspects:
\begin{itemize}
    \item \textbf{Accuracy of transition models.} Taking a fully observed initial image and its corresponding action sequence as input, we compare the difference on pixels between the image sequence calculated by the approaches and the ground truth image sequence.The MSE metric on pixels is used to test the accuracy of the learned transition models.
    \item \textbf{Planning validity and optimality.} Taking a fully observed initial image and a partially observed goal image, we invoke these approaches to compute plans and test their planning performances in terms of validity and optimality. We define the metric of planning validity rate as the proportion of valid plans on all testing instances. A plan is valid if each of its actions is valid and it finally leads to the goal state under the ground-truth transition function. Similarly, the planning optimality rate is defined as the proportion of valid plans with a length of $7$ on all testing instances. The planning cutoff time is set to 180 seconds.
   \item \textbf{Average node expansions.} To evaluate the performance of the learned heuristic model, we also 
   count the average node expansions of the approaches when computing valid plans. 
\end{itemize}


\begin{table*}[!t]
  
  \centering
  \scriptsize
  \caption{Accuracies of the learned transition models.}
  \resizebox{\textwidth}{!}{
  \begin{tabular}{c|cccc|cccc|cccc}
  \toprule
                            & \multicolumn{4}{c|}{MNIST}                & \multicolumn{4}{c|}{Mandrill}            & \multicolumn{4}{c}{Spider}               \\
                            & \aType{Rec}&\aType{Rec}$^{+}$&\aType{Lat}&\aType{Lat}$^{+}$& \aType{Rec}&\aType{Rec}$^{+}$&\aType{Lat}&\aType{Lat}$^{+}$& \aType{Rec}&\aType{Rec}$^{+}$&\aType{Lat}&\aType{Lat}$^{+}$\\
  \hline
  0 masks                   & \textbf{0.05} & \textbf{0.05} & 0.11  & 0.11     & \textbf{5.79} &\textbf{5.79} & 7.60 & 7.60     & \textbf{7.13} & \textbf{7.13} & 7.71 & 7.71      \\
  10 3$\times$3 masks       & \textbf{0.00} & 0.12    & 0.42  & 0.09      & \textbf{6.02} & 6.93   & 7.37 & 7.55     &\textbf{ 5.57}  & 6.66  & 8.18 & 6.73      \\
  20 3$\times$3 masks       &\textbf{0.00} & \textbf{0.00}    & 0.04  & 0.30       & \textbf{5.77} & 6.78   & 6.90 & 6.76      & \textbf{5.15} & 6.06   & 7.40 & 6.83      \\
  30 3$\times$3 masks       & \textbf{0.00} &\textbf{0.00}   & 0.01  & 0.01       & \textbf{5.87}  & 6.86  & 6.96 & 6.86      & \textbf{5.16} & 5.88   & 7.55 & 7.88       \\
  40 3$\times$3 masks       & \textbf{0.00}& 0.01    & 0.23  & 0.20      & \textbf{5.94}  & 6.52   & 6.91 & 6.80     & \textbf{5.31}  & 6.57   & 7.39 & 7.67     \\
  50 3$\times$3 masks       &\textbf{0.00}& 0.01    & 0.03  & 0.12     & \textbf{5.60}  & 6.86   & 6.97 & 7.13    & 6.67  & \textbf{6.01}  & 6.74 & 7.17       \\ 
  \hline
  5 3$\times$3 masks        &\textbf{0.00} & 0.01    & 0.47  & 0.19       & \textbf{5.62} & 6.69    & 7.20 & 7.11      & \textbf{5.59}  & 6.13  & 7.81 & 7.29       \\
  5 6$\times$6 masks        &\textbf{0.00} & 0.07    & 0.28  & 0.38      & \textbf{5.54}  & 6.76  & 6.77 & 7.06       & \textbf{5.96}  & 6.79  & 9.18 & 8.44       \\
  5 9$\times$9 masks        & \textbf{0.01} & \textbf{0.01}    & 0.07  & 0.32      & \textbf{6.27}  & 6.72  & 6.81 & 6.95      & \textbf{6.59}  & 8.06  & 8.08 & 6.51       \\
  5 12$\times$12 masks      & \textbf{0.01} & 0.88    & 20.27 & 4.30      & \textbf{7.21}  & 7.22  & 7.47 &  7.52   & \textbf{7.95}  & 8.26   & 9.97 & 8.55      \\ 
  \hline
  mean                      & \textbf{0.01} & 0.12   & 2.19  & 0.54       & \textbf{5.96} & 6.71   & 7.10 & 7.13      & \textbf{6.11}  & 6.76  & 8.00 & 7.48       \\
  \bottomrule
   \end{tabular}
  \label{table:transition}
    }
  \flushleft{
  For space limitation, three decimal places are missing each value should multiply by $10^{-3}$. Every row header indicates a setting of masks, the number and the size of masks. For example, ``10 3$\times$3 masks'' means each matrix contains ten pieces of 3$\times$3 masks.}
   \end{table*}

\subsection{Experimental Results}
\subsubsection{Accuracy of transition models}
{\color{black} We train the approaches on the data sets with different settings of masks and test the learned transition models on the initial image and action sequence of the testing instances. 
Table \ref{table:transition} shows the results on the transition model accuracy of the approaches. 
Compared with \aType{Lat} and its variant, \aType{Rec} and its variant \aType{Rec}$^+$ perform better.
In all domains, \aType{Rec} obtains the best performance while in the MNIST domain, it almost recovers the whole image sequences.
In average, \aType{Rec} performs 16\% better than \aType{Lat} in Mandrill and 24\% better in Spider.
Despite MNIST, the difference between \aType{Rec} and \aType{Lat} in the accuracy of the transition model is statistically significant with p-value 8.1e-5 by a two-tailed t-test.
As the p-value is far less than 0.05, it shows that \aType{Rec} improves \aType{Lat} significantly.
Furthermore, \aType{Rec} outperforms \aType{Lat}$^+$ by 16\% in Mandrill and by 18\% in Spider.
The difference between \aType{Rec} and \aType{Lat}$^+$ in the accuracy is statistically significant with p-value 7.6e-4 by a two-tailed t-test, demonstrating the significant advance of \aType{Rec}.

On the other hand, in average \aType{Rec}$^+$ beats \aType{Lat} by 39\% and \aType{Lat}$^+$ by 31\%.
By two-tailed t-tests, \aType{Rec}$^+$ differs significantly from \aType{Lat} in the transition model accuracy with p-value 0.024 and from \aType{Lat}$^+$ with p-value 0.039.
The reason why the \aType{Rec} approaches outperform the \aType{Lat} approaches is that the recurrent framework of \aType{Rec} takes a whole sequence into account, which makes up for the unknown information caused by the masks.

Surprisingly, it is counter-intuitive that the inpainting approach has a detrimental influence on the performance of \aType{Rec}.
It is because the learned model predicts incorrectly when complementing some masked regions, which leads wrong inpainting.
Such incorrect information finally makes \aType{Rec} learn a less accurate transition model.
On the other hand, 
it turns out that the inpainting approach improves \aType{Lat} via making up for the masks.
Because \aType{Lat} is not designed for the partial observed environment, the inpainting approach mitigates the consequence by the masks.
It is notable that for \aType{Lat}, the prediction model takes effect more signally on the environments with fewer masks. }

%

\subsubsection{Planning validity with different numbers and sizes of masks}
{\color{black}
To investigate how the account of small masks and the size of masks influence the planning performance, we test the validity rates on different settings of masks.
Table \ref{table:validity_rate} shows the results of the validity rates. Note that the goal image observation is also masked identically as the training sequences. 
That is, if a transition model is learned in a setting of masks than it would be tested on an instance with a goal image observation with the same setting of masks.

Unsurprisingly, our approaches \aType{Rec} and \aType{Rec}$^+$ have much better performance and solve most of the testing instances. The difference between \aType{Rec} and \aType{Lat} in the validity rate is satistically significant with p-value 5.5e-5 by a two-tailed t-test. Also, \aType{Rec}$^+$ differs significantly from \aType{Lat}$^+$ in statistics with p-value 2.1e-7 by a two-tailed t-test.
In most of the failure cases, the \aType{Lat} approaches cannot output an action sequence within the cutoff time. It is because the accuracy of its transition model is insufficient and it cannot terminate for finding an image consistent with the goal image observation.

\begin{table*}[!t]
  \centering
  \scriptsize
  \caption{The validity rates trained with images with different mask settings}
  \resizebox{\textwidth}{!}{
      \begin{tabular}{c|cccc|cccc|cccc}
      \toprule
            & \multicolumn{4}{c|}{MNIST}    & \multicolumn{4}{c|}{Mandrill} & \multicolumn{4}{c}{Spider} \\
            & \aType{Rec} & \aType{Rec}$^+$ & \aType{Lat} & \aType{Lat}$^+$ & \aType{Rec} & \aType{Rec}$^+$ & \aType{Lat} & \aType{Lat}$^+$ & \aType{Rec} & \aType{Rec}$^+$ & \aType{Lat} & \aType{Lat}$^+$ \\
      \hline
      0 masks & \textbf{98}    & \textbf{98}    & 70    & 70    & \textbf{95}    & \textbf{95}    & 71    & 71    & \textbf{98}    & \textbf{98}    & 62    & 62 \\
       10 3$\times$3 masks & \textbf{98}    & \textbf{98}    & 64    & 76    &  \textbf{98}    & \textbf{98}   & 73    & 83    & 95    & \textbf{96}    & 67    & 66 \\
       20 3$\times$3 masks & \textbf{99}    & \textbf{99}    & 80    & 76    &  \textbf{99}    & 98    & 69    & 76    & 95    & \textbf{99}    & 69    & 78 \\
       30 3$\times$3 masks & \textbf{100}   & \textbf{100}   & 76    & 68    & \textbf{97}    & 95    & 72    & 68    & 96    & \textbf{100}   & 75    & 67 \\
       40 3$\times$3 masks &\textbf{98}    & \textbf{98}    & 66    & 72    & 95    & \textbf{96}    & 80    & 74    & 97    & \textbf{99}    & 77    & 74 \\
       50 3$\times$3 masks & \textbf{98}    & 94    & 75    & 72    & \textbf{99}    & 95    & 72    & 77    & \textbf{96}    & 95    & 72    & 72 \\
      \hline
       5 3$\times$3 masks & \textbf{97}    & 96    & 78    & 75    & \textbf{96}    & \textbf{96}    & 74    & 81    & 98    & \textbf{99}    & 66    & 73 \\
       5 6$\times$6 masks & \textbf{97}    & 94    & 66    & 75    & \textbf{97}    & \textbf{97}    & 72    & 75    & \textbf{97}    & 95    & 58    & 69 \\
       5 9$\times$9 masks & \textbf{100}   & 98    & 56    & 67    & \textbf{98}    & 95    & 77    & 71    & \textbf{98}    & \textbf{98}    & 78    & 80 \\
       5 12$\times$12 masks & \textbf{95}    & 91    & 19    & 64    & 93    & \textbf{96}    & 77    & 81    & \textbf{99}    & 98    & 74    & 82 \\
      \hline
      mean  & 98    & 96.6  & 65    & 71.5  & 96.7  & 96.1  & 73.7  & 75.7  & 96.9  & 97.7  & 69.8  & 72.3 \\
      \bottomrule
      \end{tabular}%
  \label{table:validity_rate}
  }%
  \flushleft{
  The percentage symbol ``\%'' is not shown.}
   \end{table*}

Essentially, the results of their validity rates conform to the results of their transition accuracies. 
The validity rates of \aType{Rec} and \aType{Rec}$^+$ are comparable. While the validity of \aType{Lat}$^+$ is higher than that of \aType{Lat}. In some settings, though our inpainting approach does not improve \aType{Lat} in the transition model accuracy but improves its performance in planning because it makes the latent state of the image observation closer to that of the ground-turth image and it is easier to find an appropriate action towards to the goal.
}


%

\begin{table*}[!th]
\centering
\scriptsize
\caption{Optimality rates trained with images with different masks}
\resizebox{\textwidth}{!}{
    \begin{tabular}{c|cccc|cccc|cccc}
    \toprule
          & \multicolumn{4}{c|}{MNIST}    & \multicolumn{4}{c|}{Mandrill} & \multicolumn{4}{c}{Spider} \\
          & \aType{Rec} & \aType{Rec}$^+$ & \aType{Lat} & \aType{Lat}$^+$ & \aType{Rec} & \aType{Rec}$^+$ & \aType{Lat} & \aType{Lat}$^+$ & \aType{Rec} & \aType{Rec}$^+$ & \aType{Lat} & \aType{Lat}$^+$ \\
    \hline
  0 masks & \textbf{98}    & \textbf{98}   & 68    & 68    & 95    & 95    & 70    & 70    & 98    & 98    & 57    & 57 \\
    
    10 3$\times$3 masks & \textbf{98}    & \textbf{98}   & 60    & 74    & \textbf{98}    & \textbf{98}    & 72    & 79    & 95    & \textbf{96}    & 66    & 66 \\
    
    20 3$\times$3 masks  & \textbf{99}    & \textbf{99}  & 76    & 74    & \textbf{98}    & \textbf{98}    & 65    & 75    & 95    & \textbf{98}    & 68    & 77 \\
    
    30 3$\times$3 masks & \textbf{100}   & \textbf{100}   & 74    & 66    & \textbf{97}    & 95    & 68    & 68    & 96    & \textbf{100}   & 73    & 64 \\
    
    40 3$\times$3 masks & \textbf{98}    & \textbf{98}    & 64    & 70    & 95    & \textbf{96}    & 77    & 73    & 97    & \textbf{99}    & 75    & 68 \\
    
    50 3$\times$3 masks & \textbf{98}    & 94    & 73    & 68    & \textbf{99}    & 95    & 70    & 73    & \textbf{96}    & 95    & 69    & 68 \\
   \hline 
    5 3$\times$3 masks & \textbf{97}    & 96    & 73    & 73    & \textbf{96}    & \textbf{96}    & 71    & 78    & 98    & \textbf{99}    & 59    & 70 \\
    
    5 6$\times$6 masks & \textbf{97}    & 94    & 65    & 73    & \textbf{97}    & \textbf{97}    & 70    & 69    & \textbf{97}    & 95    & 50    & 68 \\
    
    5 9$\times$9 masks & \textbf{100}   & 98    & 56    & 66    & \textbf{98}    & 95    & 72    & 66    & 97    & \textbf{98}    & 78    & 79 \\
    
    5 12$\times$12 masks & \textbf{95}    & 91    & 12    & 60    & 93    & \textbf{96}    & 75    & 77    & \textbf{99}    & 98    & 65    & 71 \\
    \hline
    mean & 98	&96.6	&62.1	&69.2	&96.6	&96.1	&71	&72.8	&96.8	&97.6	&66	&68.8\\
    \bottomrule
    \end{tabular}%
\label{table:optimality_rate}
}
\flushleft{
The percentage symbol ``\%'' is not shown. }
 \end{table*}

 \subsubsection{Planning optimality with different sizes of masks}
{\color{black}
We also evaluate the influence of masks on the planning optimality performance of the approaches.
Table \ref{table:optimality_rate} shows the results of the optimality rates under different numbers and sizes of masks. 
It is not difficult to find that \aType{Rec} and \aType{Rec}$^+$ are superior to \aType{Lat} and \aType{Lat}$^+$ in the optimality rate.
In average, \aType{Rec} outperforms \aType{Lat} by up to 47\% and \aType{Lat}$^+$ by up to 38\%.
For statistically significance analysis, the difference between \aType{Rec} and \aType{Lat} has a p-value 4.4e-5 by a two-tailed t-test.
With our inpainting approach, \aType{Rec}$^+$ is able to optimally solve 38\% more instances than \aType{Lat}$^+$.
The difference between \aType{Rec}$^+$ and \aType{Lat}$^+$ is statistically significantly with p-value 7.1e-8 by a two-tailed t-test.

Surprisingly, most of valid plans computed by the \aType{Rec} approaches are optimal.
For only one instance in Mandrill and one in Spider, \aType{Rec} cannot compute a valid and non-optimal plan which contains repeated actions.
That is, the ratio of its optimality rate on its validity rate averagely is extremely closed to 100\%.
Whereas, the optimality rates of the \aType{Lat} approaches is lower than their validity rates.
Considering the \aType{Lat} approaches use a goal-distance heuristics function,
the superior results of the \aType{Rec} approaches benefit from the excellent guiding performance of the learned heuristic model on selecting actions. 
}

\begin{table*}[!t]
\centering
\scriptsize
\caption{Average node expansions of finding valid plans}
\resizebox{\textwidth}{!}{
    \begin{tabular}{c|cccc|cccc|cccc}
    \toprule
          & \multicolumn{4}{c|}{MNIST}    & \multicolumn{4}{c|}{Mandrill} & \multicolumn{4}{c}{Spider} \\
          & \aType{Rec} & \aType{Rec}$^+$ & \aType{Lat} & \aType{Lat}$^+$ & \aType{Rec} & \aType{Rec}$^+$ & \aType{Lat} & \aType{Lat}$^+$ & \aType{Rec} & \aType{Rec}$^+$ & \aType{Lat} & \aType{Lat}$^+$ \\
    \hline
      0 masks & 42.57 & 42.57 & 3848.46 & 3848.46 & 31.49 & 31.49 & 4271.66 & 4271.66 & 48.94 & 48.94 & 2891.35 & 2891.35 \\
      10 3$\times$3 masks & 40.49 & 44.90  & 4280.08 & 4407.42 & 34.45 & 38.45 & 5459.95 & 3212.90 & 43.34 & 39.63 & 4067.67 & 2235.79 \\
      20 3$\times$3 masks & 40.69 & 36.69 & 2633.68 & 5513.58 & 33.66 & 34.00    & 4679.3 & 2302.59 & 40.38 & 35.92 & 3513.22 & 2126.11 \\
      30 3$\times$3 masks & 39.40  & 35.68 & 2949.11 & 2277.23 & 37.57 & 40.84 & 5140.56 & 1907.06 & 36.83 & 36.56 & 3746.44 & 3242.48 \\
      40 3$\times$3 masks & 40.61 & 33.55 & 2032.91 & 1861.8 & 38.06 & 47.92 & 3034.20 & 2777.97 & 39.42 & 34.02 & 3352.30 & 1784.29 \\
      50 3$\times$3 masks & 34.45 & 35.28 & 2073.00  & 2119.77 & 44.65 & 41.77 & 4124.56 & 2404.91 & 37.46 & 33.73 & 4741.21 & 2190.90 \\
      \hline 
      5 3$\times$3 masks & 34.31 & 34.54 & 3096.56 & 3921.97 & 35.54 & 39.38 & 3104.75 & 4302.73 & 42.24 & 39.27 & 6868.38 & 2385.33 \\
      5 6$\times$6 masks & 35.92 & 33.85 & 4734.08 & 3889.55 & 38.31 & 41.36 & 2671.25 & 2734.22 & 35.42 & 33.68 & 4518.10 & 2975.04 \\
      5 9$\times$9 masks & 41.16 & 41.47 & 1492.07 & 2465.31 & 34.90  & 35.20  & 1612.52 & 3125.46 & 40.24 & 40.57 & 1735.28 & 1059.40 \\
      5 12$\times$12 masks & 54.02 & 43.52 & 3805.05 & 5427.38 & 40.26 & 36.38 & 2850.86 & 1929.88 & 39.60  & 43.84 & 3896.27 & 3513.61 \\
      \hline
      mean & 38.38  & 37.00  & 2911.44  & 3307.08  & 37.14  & 39.87  & 3728.39  & 2845.98  & 39.42  & 36.67  & 4067.83  & 2249.92  \\
    \bottomrule
    \end{tabular}%
\label{table:searching_node}
}%
 \end{table*}

\subsubsection{Planning efficiency}
{\color{black}
We also evaluate the searching ability of the approaches by counting the average node expansions for computing valid plans.
The results are depicted in 
Table \ref{table:searching_node}. 
To find a valid plan, the \aType{Lat} approaches need to expand two orders of magnitude more nodes than the \aType{Rec} approaches. 
By a two-tailed t-test, the difference between \aType{Rec} and \aType{Lat} is statistically significant with p-value 6.9e-6.
Likewise, the difference between \aType{Rec}$^+$ and \aType{Lat}$^+$ is statistically significant with p-value 6.8e-6.

Considering that most of the valid plans by \aType{Rec} and \aType{Rec}$^+$ are optimal, each of which contains seven steps, the searching procedure based on the learned heuristic model expands about five nodes averagely at each step.
It also shows that the learned heuristic model is able to return an appropriate action nearly at every step.
}

\ignore{
\subsection{Case Study}
\begin{figure}[!ht]
    \centering
    \includegraphics[width=0.5\textwidth]{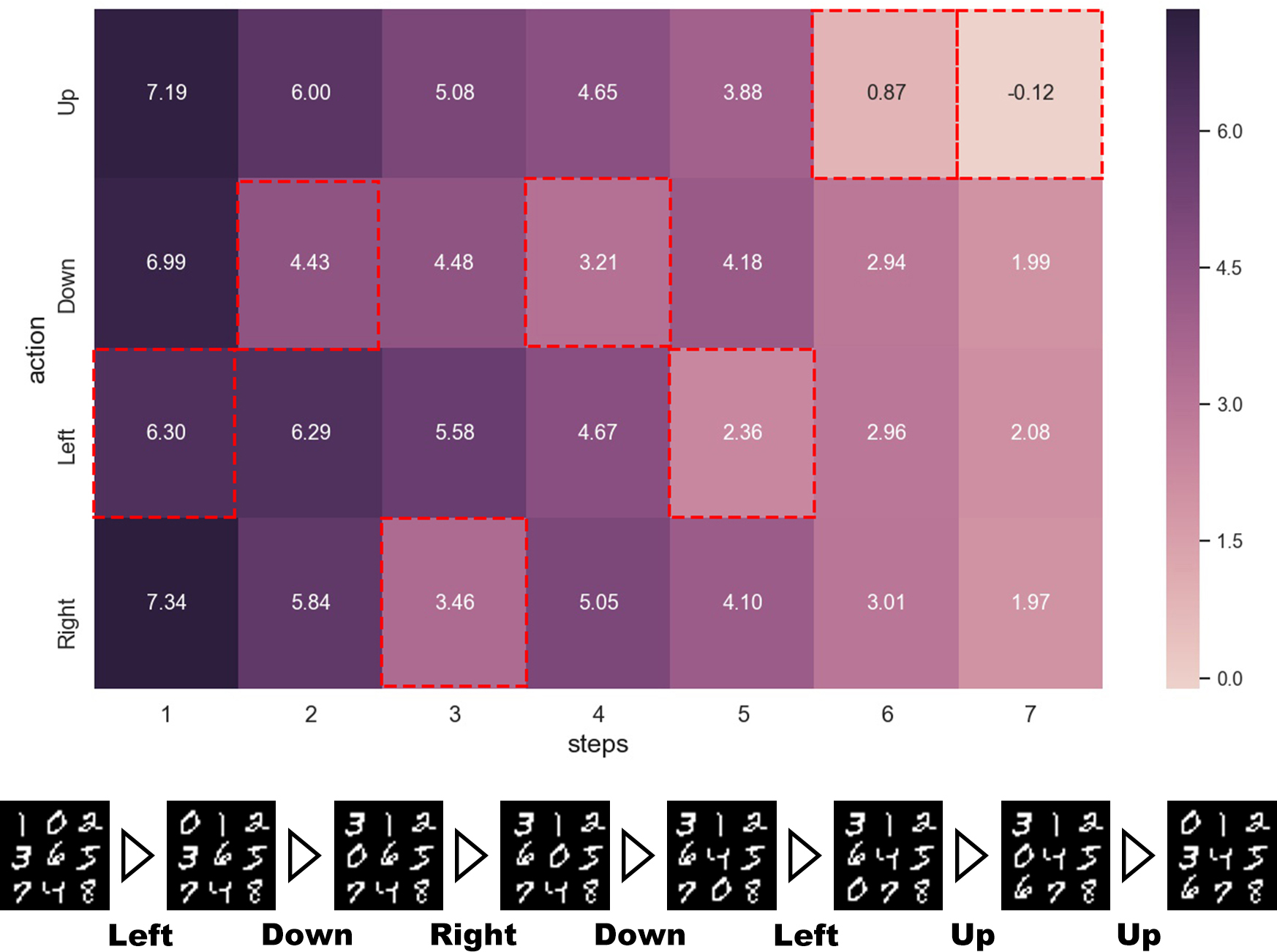}
    \caption{Heuristic}
    \label{heuristic}
\end{figure}
}

\section{Conclusion}
A growing number of people are interested in learning techniques, especially those based on unstructured data, due to its availability and commonality. In this paper, we focus on planning model learning on images, which can be easily obtained from cameras.
Comparing to the assumption that every image is fully observed, we tackle a more realistic scenario where every image is partially observed. 
More specifically, we present a novel visual planning model learning framework that is applicable to an environment with incomplete observations.
By conducting experiments on the three domains, we show the superiority of our approach on transition model learning and planning performance.

In this paper, we do not learn the condition where an action is valid, which is generally a part of a planning model.
It is in fact difficult in a partially observable environment. If the learned precondition is too strong, it probably happens that there would be no action to be applicable; if it is too weak, an invalid action may be chosen.  
Generally, learning a sound and complete model about action precondition requires a highly accurate prediction on the masked regions. We take the task of learning precondition as one of our future works.

\section{Acknowledgement}
This research was funded by the National Natural Science Foundation of China (Grant
No. 62076263, 61906216, 61976232), Guangdong Natural Science Funds for Distinguished Young Scholar (Grant No. 2017A030306028), Guangdong Special Branch Plans Young Talent with Scientific and Technological Innovation (Grant No. 2017TQ04X866), Guangdong Basic and Applied Basic Research Foundation (No. 2022A1515011355, 2020A1515010642), Pearl River Science and Technology New Star of Guangzhou and Guangdong Province Key Laboratory of Big Data Analysis and Processing, Guizhou Science Support Project (No. 2022-259), Humanities and Social Science Research Project of Ministry of Education (18YJCZH006).
 \bibliographystyle{apalike} 
 \bibliography{elsarticle-template-num}





\end{document}